# KG-FGNN: Knowledge-guided GNN Foundation Model for Fertilisation-oriented Soil GHG Flux Prediction

Yu Zhang[1], Gaoshan Bi[1], Simon Jeffery[2], Max Davis[2], Yang Li[3], Qing Xue[3], Po Yang[1*]

*Abstract*—Precision soil greenhouse gas (GHG) flux prediction is essential in agricultural systems for assessing environmental impacts, developing emission mitigation strategies and promoting sustainable agriculture. Due to the lack of advanced sensor and network technologies on majority of farms, there are challenges in obtaining comprehensive and diverse agricultural data. As a result, the scarcity of agricultural data seriously obstructs the application of machine learning approaches in precision soil GHG flux prediction. This research proposes a knowledge-guided graph neural network framework that addresses the above challenges by integrating knowledge embedded in an agricultural process-based model and graph neural network techniques. Specifically, we utilise the agricultural process-based model to simulate and generate multi-dimensional agricultural datasets for 47 countries that cover a wide range of agricultural variables. To extract key agricultural features and integrate correlations among agricultural features in the prediction process, we propose a machine learning framework that integrates the autoencoder and multi-target multi-graph based graph neural networks, which utilises the autoencoder to selectively extract significant agricultural features from the agricultural process-based model simulation data and the graph neural network to integrate correlations among agricultural features for accurately predict fertilisation-oriented soil GHG fluxes. Comprehensive experiments were conducted with both the agricultural simulation dataset and real-world agricultural dataset to evaluate the proposed approach in comparison with well-known baseline and state-of-the-art regression methods. The results demonstrate that our proposed approach provides superior accuracy and stability in fertilisation-oriented soil GHG prediction.

## I. INTRODUCTION

THE challenges in the field of agriculture are crucial for humanity. As the global population continues to grow, approximately 780 million people are currently facing hunger, and it is projected that the population will reach 9 billion by 2050, with a 60 per cent increase in the demand for food [1]. In order to enhance yields, agricultural systems rely heavily on chemical fertiliser inputs, but the resulting environmental problems are becoming increasingly serious, especially soil greenhouse gas (GHG) (including $N_2O$ and $CO_2$) emissions caused by fertilisation practices, which can exacerbate climate change, threaten ecosystems and human health [2]. Therefore, it is significant to construct an effective approach for predicting soil GHG fluxes to develop precision fertiliser application strategies and promote green transformation of agriculture. Artificial Intelligence (AI) technologies have been progressively applied to the agricultural field in recent years, including automated machinery [3][4], pest and disease monitoring [5][6], fertiliser application prediction [7][8] and environmental early warning systems [9][10], which have shown great potential in enhancing efficiency and reducing resource waste. However, the application of machine learning to soil GHG flux prediction remains challenging, especially in the context of the lack of structured agricultural data, the complexity and variability of fertiliser application practices.

The response of soil systems to agricultural management practices including fertiliser application is highly complex and non-linear. Modern agricultural technologies have provided a variety of innovative management strategies, including intelligent fertiliser application to improve yields while reducing environmental impacts. Advanced AI approaches can assist in determining optimal management parameters to improve prediction accuracy, avoid resource wastage and environmental pollution. Traditional methods for estimating GHG fluxes mainly base on static models or field measurements, including empirical emission factor methods and mechanistic models, which typically require a large number of experiments, with complex model parameters, limited applicability, and large uncertainties in different regions, soil types, and management scenarios [11][12]. In addition, obtaining high-quality GHG emission data remains a serious challenge. Although national and international agricultural research organisations have accumulated a certain amount of GHG-related data resources, which can theoretically provide support for machine learning approaches, these data suffer from the defects of being irrecoverable, difficult to interpret and lacking in generality [13]. Incomplete, biased or irrelevant data can significantly affect model performance and can erode the trust of farmers and agricultural practitioners in digital environmental management tools, which in turn affects the development of climate-smart agriculture. To the best of our knowledge, there is no recognised, publicly available database of soil GHG emissions. More critically, data-driven prediction approaches constructed on the correlation between fertiliser application practices and GHG responses are still a vacant area that lacks systematic research.

[1] School of Computer Science, University of Sheffield, Sheffield, UK
[2] Agriculture and Environment, Harper Adams University, Newport, UK
[3] Research and Development, Mutus Tech Ltd, Liverpool, UK
* Corresponding author: Po Yang (po.yang@sheffield.ac.uk)

The aim of this research is to construct a fertilisation-oriented machine learning foundation model for predicting soil GHG fluxes, combined with multi-dimensional agricultural simulation data generated by the agricultural process-based model DSSAT [14], to enhance the prediction accuracy and stability of various soil GHG fluxes under data-scarce conditions. However, the design and implementation of the approach faced multiple challenges. Firstly, it is complex and costly to collect and standardise multi-source agricultural data on soil, climate, cropping and fertilisation from various countries, regions and farms to construct comprehensive datasets that can be utilised to train the machine learning approach. Secondly, the seamless integration of domain knowledge embedded in agricultural process-based model and agricultural feature correlations in machine learning models is a key challenge for constructing efficient approaches. Thirdly, the difficulty in algorithm design exists in how to logically construct the network structure and achieve multi-target joint optimisation in facing the multi-dimensional heterogeneous correlations among agricultural variables.

To address above challenges, we constructed a multi-dimensional agricultural simulation dataset with the agricultural process-based model and combined it with real-world farm data for model construction, training and testing. The agricultural data contains key agricultural variables including climate information, soil nutrients, crop yields, fertiliser application records, and soil GHG emissions. The multi-source heterogeneous agricultural data provide a solid data foundation for modelling the coupled interaction between fertiliser application practices and soil GHG emissions. In terms of algorithms, this research proposes a knowledge-guided GNN regression framework (Fig. 1) that combines a modified autoencoder with a multi-target multi-graph structure to enhance the ability to predict various types of soil GHG fluxes at multiple time points. Specifically, we first utilise the agricultural process-based model DSSAT to generate agricultural multi-dimensional variable data covering 47 countries as a source of fundamental features; then we utilise a modified autoencoder to selectively extract significant agricultural variables from the agricultural simulation dataset to reduce the dimensionality, decouple, and retain key information; finally, a multi-target multi-graph GNN mechanism was designed and constructed to model the structured correlations among agricultural variables to achieve precise prediction of fertilisation-oriented soil GHG fluxes.

The main contributions of this research are as follows:
- We utilised the agricultural process model to generate and construct a multi-dimensional agricultural simulation dataset covering 47 countries, which provides a solid data base for fertilisation-oriented soil GHG flux prediction.
- We propose a novel GNN framework integrating a modified autoencoder with a multi-target multi-graph structure for extracting key agricultural features from the simulation data and modelling structured correlations between agricultural variables to enhance the accuracy and stability of soil GHG flux predictions under various fertiliser application scenarios.
- Extensive experiments are conducted with genuine farm data to evaluate and demonstrate the feasibility and effectiveness of our proposed approach in real-world scenarios.

## II. RELATED WORK

Precision agriculture is a system that integrates modern agricultural management methods scientifically at a specific time, place and quantity based on geographic variability and advanced information technology [15]. Its core concept is to precisely deploy agricultural resources based on climate, soil and crop conditions under the premise of protecting the environment, in order to maximise agricultural output and optimise resource utilisation [16]. Soil GHG flux monitoring and prediction are essential for assessing the environmental impacts of agricultural practices, optimising fertiliser management strategies, and achieving sustainable agriculture as a primary component of precision agriculture. Fertilisation-oriented soil GHG prediction technology aims to recognize the trends of GHG fluxes of $N_2O$, $CO_2$ and $CH_4$ released from soils with various fertilisation practices based on the fertiliser requirement pattern of crops, the nutrient status of soils, and the behaviour of fertiliser application [17]. This process is related to fertiliser application efficiency and crop yield, which directly affects the total agricultural GHG emissions and ecological safety [18]. By precisely modelling the dynamic correlation between soil nutrients and GHG emissions, scientific agricultural GHG monitoring and regulation can be achieved to promote sustainable and low-emission agricultural development. To achieve fertilisation-oriented soil GHG flux prediction, it involves comprehensive integration of multi-source variables including fertilisation practices, soil properties and climatic factors [19]; construction of interpretable prediction approaches to model and analyse the coupling mechanism between fertilisation decisions and soil GHG responses [20]; and implementation of targeted regulatory measures in practical production to achieve the goal of reducing emissions and increasing efficiency [2].

In the field of machine learning, various algorithms and models have been proposed for modelling and predicting agricultural GHG emissions. Neural network-based methods have been applied to the task of estimating soil carbon emissions and GHG fluxes in specific regions and crop types [17]. Traditional machine learning algorithms including support vector machines and random forests have been utilised to model the GHG emission process, typically integrating input variables including crop cultivation, climatic factors and soil information [21]. Partial research has attempted to integrate agricultural knowledge graphs with machine learning techniques to construct a soil carbon emission assessment system, which analyses and predicts emission trends with historical observation data and soil nutrient information [22][23]. However, there are numerous limitations in the existing research. Firstly, the majority of approaches are unable to adequately explore the complex interaction between spatio-temporal features and multi-dimensional variables contained in agricultural data, which makes it difficult to effectively enhance the model performance. Second, existing research lacks systematic modelling of the correlation mechanism between fertiliser application and soil GHG fluxes, which can be difficult to be utilised for guiding the emission reduction strategies with practical fertiliser management. Thirdly, agricultural data have significant scarcity, incompleteness and heterogeneity, and the existing models generally exhibit weak robustness and generalisation ability when processing these data, which are difficult to achieve the urgent requirements for high-precision prediction of soil GHG emissions in the context of precision agriculture and climate-smart agriculture.

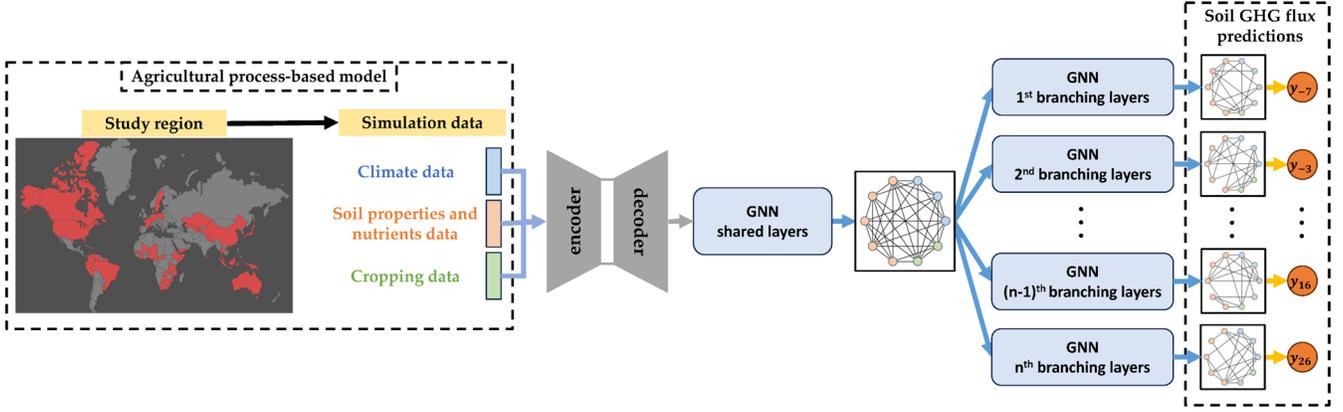

**Fig. 1.** Overview of the method and framework used for proposed KG-FGNN. The proposed approach filters and decouples domain knowledge from agricultural process-based models and utilises a multi-target multi-graph architecture incorporating shared and branching structures to enhance the accuracy, stability and interpretability of fertilisation-oriented soil GHG flux multi-target predictions, while maintaining global generalisation capabilities.

To address the above challenges, we constructed an agricultural multi-dimensional simulation dataset covering 47 countries with the agricultural process-based model and combined it with real-world farm data for model construction, training and testing. Targeting the highly complex structured correlations among agricultural variables, we propose a knowledge-guided GNN regression framework to predict various types of soil GHG fluxes at multiple time points. GNNs have demonstrated powerful capabilities in handling non-Euclidean structured data, and are particularly applicable to modelling complex spatial and logical interactions between agricultural variables. GNNs have been applied to tasks including crop identification, yield prediction, and meteorological data modelling with promising results in capturing the interaction features among variables [24][25][26]. Meanwhile, knowledge-guided machine learning models are gaining growing attention, especially in the fields of agriculture and environmental sciences. By combining domain knowledge in agricultural process-based models with data-driven models, the model's interpretability and generalisation ability can be effectively enhanced [22][27]. In this research, we further proposed a multi-graph structure to express the heterogeneous correlations among diverse agricultural variables and integrated with a modified autoencoder to filter and decouple the agricultural input features, which enhanced the ability to model the coupled relationship between fertiliser application and soil GHG fluxes. This approach enhances the prediction accuracy and stability of soil GHG fluxes in complex agricultural scenarios, and provides a novel path for the integration of agricultural process-based knowledge with graph modelling.

### III. METHODOLOGY

*A. Problem formulation*

Let $\mathcal{D} = \{(\mathbf{x}^{(i)}, \mathbf{y}^{(i)})\}_{i=1}^{N}$ denote a collection of $N$ observations obtained from field-scale winter-wheat trials. For each instance, the input feature vector $\mathbf{x}^{(i)} \in \mathbb{R}^d$ (with $d = 10$) summarises fertiliser management, climate information, and soil properties. The target $\mathbf{y}^{(i)} \in \mathbb{R}^T$ (with $T = 9$) contains $N_2O$ or $CO_2$ fluxes measured at nine discrete time offsets following application (−7, -3, 0, 1, 2, 4, 8, 16, 26 day, day 0 is the fertiliser application day). Our goal is to learn a mapping $f: \mathbb{R}^d \to \mathbb{R}^T, \mathbf{x} \mapsto \hat{\mathbf{y}} = f(\mathbf{x})$, minimising the expected squared regression error whilst simultaneously preserving the intrinsic covariance structure of $\mathbf{x}$. We therefore optimise the following composite loss:

$$\mathcal{L} = \frac{1}{N}\sum_{i=1}^{N}\left(\left\|f(\mathbf{x}^{(i)}) - \mathbf{y}^{(i)}\right\|_2^2 + \alpha\left\|g(\mathbf{x}^{(i)}) - \mathbf{x}^{(i)}\right\|_2^2\right) \quad (1)$$

where $g$ denotes an autoencoder decoder and $\alpha \in [0,1]$ balances supervised and self-supervised terms.

*B. Knowledge-guided autoencoder-GNN architecture*

   ***Autoencoder feature embedding***

In typical agricultural scenarios, the available meteorological, soil nutrient and fertiliser management data are generally highly coupled, scale heterogeneous and seriously lacking, directly inputting them into the prediction model will amplify the noise and lead to overfitting, difficult to retain the essential nonlinear features of the fertiliser application drive. In this research, we firstly utilise the modified autoencoder technique to compress and de-redundancy the original variables, thus highlighting the intrinsic information of fertiliser-environment interactions in the low-dimensional latent space, and laying a solid foundation for the subsequent multi-graph GNN to construct structured agricultural correlation knowledge graphs. A multilayer perceptron encodes $\mathbf{x}$ into a latent representation $\mathbf{z} \in \mathbb{R}^h$:

$$\mathbf{z} = \phi_E(\mathbf{x}) = \sigma_L(\mathbf{W}_L \sigma_{L-1}(\ldots \sigma_1(\mathbf{W}_1 \mathbf{x} + \mathbf{b}_1) \ldots) + \mathbf{b}_L) \quad (2)$$

with ReLU activations $\sigma_\ell$ and learnable weights $\{\mathbf{W}_\ell, \mathbf{b}_\ell\}$. A mirror-symmetric decoder $g = \phi_D$ reconstructs $\hat{\mathbf{x}} \in \mathbb{R}^d$.

TABLE I. THE STRUCTURE OF WINTER WHEAT AGRICULTURAL DATASET.

| Agricultural feature | | | Winter wheat dataset target | |
|---|---|---|---|---|
| **Climate data** | **Soil properties and nutrients data** | **Cropping data** | **Soil GHG flux** | **Range of target values** |
| Mean daily temperature (°C) Monthly rainfall (mm) Monthly solar radiation (TJ/ha) | Soil pH value Soil water holding capacity (mm) Soil phosphorus content (mg/l) Soil potassium content (mg/l) Soil magnesium content (mg/l) | N fertiliser applied (kg/ha) Seeds sown per m$^2$ | $N_2O$ $CO_2$ | 0 – 10348.62 (μg $N_2$O-N/h/m$^2$) 0 – 456.17 (mg $CO_2$-C/h/m$^2$) |

*Graph representation and node characterisation*

In order to adequately portray the complex and non-linear interactions between climate, soil and fertiliser management variables in authentic agricultural scenarios, we re-expressed the low-dimensional latent feature space as a graph structure. The graph representation was motivated by two practical considerations: 1) there are heterogeneous and non-Euclidean dependencies between the agricultural variables, which are difficult to express explicitly in a traditional tabular model; and 2) there are significant differences between farms in terms of climatic zones, soil types, and management modes, explicit relational modelling allows for injecting agronomical domain knowledge into the graph edges. The graphical skeleton enables the model to simultaneously investigate direct and higher-order interactions, which enhances the robustness and interpretability of fertilisation-oriented soil GHG flux predictions. Each element of $\mathbf{z}$ is linearly projected to yield node-wise embeddings:

$$\mathbf{H}^{(0)} = \text{reshape}(\mathbf{W}_p \mathbf{z} + \mathbf{b}_p) \in \mathbb{R}^{d \times 1} \quad (3)$$

where nodes correspond one-to-one with the original features.

*Graph convolution and multi-task branching*

In agricultural practices, different soil physicochemical indicators and meteorological factors can jointly influence fertilisation-oriented soil GHG emissions in a cross-cutting and heterogeneous manner; meanwhile, field management requires that the model can simultaneously predict multiple GHG fluxes in multiple temporal windows to guide dynamic fertiliser adjustment decisions. It is challenging for models to simultaneously capture these higher-order dependencies among variables and time periods if only a single regression header or planar feature interaction is utilised, thus weakening the guidance value for dynamic fertiliser regulation. In order to balance the holistic capture of general correlations and the fine-grained modelling of target-specificity in the same framework, we construct a fully connected graph on latent feature nodes and propose a shared-branching graph convolution mechanism: the shared layer utilises a spectrally normalised adjacency matrix to globally propagate the information and to learn the universal coupling patterns among fertiliser application and climate-soil variables; Each task branch is further refined for various time points to model its unique nonlinear response mechanism at a fine-grained level, which significantly enhances the multi-target prediction accuracy, stability and interpretability while maintaining the global generalisation capability. We place a fully-connected, undirected graph $G = (V, E)$ over the $d$ nodes ($|E| = 2\binom{d}{2}$). Edge weights $\mathbf{w} \in \mathbb{R}^{|E|}$ are trainable and shared across the mini-batch. A stack of $S$ shared graph-convolution layers update the node states:

$$\mathbf{H}^{(\ell+1)} = \sigma(\widetilde{\mathbf{D}}^{-1/2} \widetilde{\mathbf{A}} \widetilde{\mathbf{D}}^{-1/2} \mathbf{H}^{(\ell)} \mathbf{W}^{(\ell)}), \ell = 0, \ldots, S-1 \quad (4)$$

where $\widetilde{\mathbf{A}} = \mathbf{A} + \mathbf{I}$ and $\widetilde{\mathbf{D}}_{ii} = \sum_j \widetilde{\mathbf{A}}_{ij}$.

Distinct regression heads are then instantiated for each target $t = 1, \ldots, T$. Head $t$ comprises $B_t$ graph-convolution layers followed by a linear node-wise read-out:

$$\mathbf{O}_t = \text{GCN}_t(\mathbf{H}^{(S)}), \hat{y}_t = \frac{1}{d} \sum_{n=1}^d [\mathbf{O}_t]_n \quad (5)$$

IV. EXPERIMENTAL SETTINGS

*A. Agricultural process-based simulation dataset*

We utilised the agricultural process-based model DSSAT to simulate and generate multi-dimensional agricultural dataset for 47 countries covering various agricultural variables including climate, soil, cropping data, fertiliser application and soil GHG emissions. The DSSAT can utilise weather, soil, crop type, management, and observational experimental data to simulate responses, we simulated yield and soil GHG emission results for 224 real-world winter wheat fertilisation plans with various weather and soil conditions. The fertilisation simulation data amount has reached 319,536, the fertilisation-oriented soil GHG flux simulation data amount has reached 1,081,179, as each fertilisation plan has multiple applications. The simulation dataset contains various types of agricultural data and the factors selected must satisfy two standards. First, it can affect soil GHG emissions and crop growth from agronomic standpoints. Second, it is a value that is easily accessible in real-world farms, otherwise it can significantly affect the real-world applicability of the approach. Specifically, our dataset contains three categories. The first category is climatic data, which can be collected via weather forecasting tools and contains three factors: mean daily temperature, monthly rainfall and monthly solar radiation. The second category is soil properties and nutrient data, which can be obtained via soil tests. Soil properties consist of two factors: soil water holding capacity and soil pH, and soil nutrients consist of three factors: soil phosphorus, potassium and magnesium content. The third category is cropping data, which can be collected from cropping records and contains two factors: seed sown per $m^2$ and nitrogen fertiliser applied. Overall, there are 10 input agricultural features utilised for fertilisation-oriented soil GHG flux prediction. The structure of the winter wheat dataset is summarised in Table I.

TABLE II. EVALUATION RESULTS FOR FERTILISATION-ORIENTED SOIL $CO_2$ FLUX PREDICTION WITH AGRICULTURAL PROCESS-BASED SIMULATION WINTER WHEAT DATASET. THE BEST RESULTS ARE BOLDED.

| Regression models | rMSE | MAE |
|---|---|---|
| Ridge regression | 42.0851±1.2001 | 31.5828±1.1075 |
| Lasso regression | 79.4467±1.3797 | 64.0175±1.2206 |
| Elastic-Net | 67.5203±1.3788 | 51.3836±1.2038 |
| ResNet | 32.0351±1.7427 | 21.9816±1.5305 |
| GCN | 72.3625±1.9040 | 57.5899±1.5736 |
| GraphSAGE | 34.8408±1.5203 | 22.4711±1.4801 |
| KG-FGNN (ours) | **23.1547±0.6841** | **15.3709±0.2087** |

TABLE III. EVALUATION RESULTS FOR FERTILISATION-ORIENTED SOIL $N_2O$ FLUX PREDICTION WITH AGRICULTURAL PROCESS-BASED SIMULATION WINTER WHEAT DATASET. THE BEST RESULTS ARE BOLDED.

| Regression models | rMSE | MAE |
|---|---|---|
| Ridge regression | 73.5245±3.9411 | 11.1318±1.2393 |
| Lasso regression | 74.2990±3.9903 | 8.8350±1.2387 |
| Elastic-Net | 73.6533±3.7323 | 8.7117±1.2034 |
| ResNet | 67.9032±6.6019 | 12.0417±1.4582 |
| GCN | 69.4923±8.9385 | 8.7825±1.2506 |
| GraphSAGE | 68.9962±8.3079 | 9.8206±1.4066 |
| KG-FGNN (ours) | **46.9194±2.2419** | **6.0409±0.1975** |

*B. Evaluation metrics*

We designed and constructed a knowledge-guided GNN regression framework for predicting fertilisation-oriented soil GHG fluxes. In this research, root mean square error (rMSE) and mean absolute error (MAE) are utilised as main metrics for assessing the performance of various prediction algorithms. R-squared ($R^2$) tells us how well the regression model can predict the value of response variable in terms of percentage and is utilised to assess the fit of predicted values to the actual values; $R^2$ can range from $-\infty$ to 1, with the closer the value is to 1, the better the prediction performance. rMSE, MAE and $R^2$ are defined as follows:

$$\text{rMSE}(y, \hat{y}) = \sqrt{\frac{\|y-\hat{y}\|_2^2}{n}} \qquad (6)$$

$$\text{MAE}(y, \hat{y}) = \frac{1}{n}\sum_{i=1}^{n} |y_i - \hat{y}_i| \qquad (7)$$

$$R^2 = 1 - \frac{\sum_{i=1}^{n}(y_i - \hat{y}_i)^2}{\sum_{i=1}^{n}(y_i - \bar{y})^2} \qquad (8)$$

where $y$ is the vector of true target values, $\hat{y}$ is the vector of model-predicted values, $n$ denotes the number of samples, and $\|\cdot\|_2$ is the $\ell_2$ norm. For the point-wise forms, $y_i$ and $\hat{y}_i$ are the true and predicted value for the $i^{\text{th}}$ sample respectively, while $\bar{y}$ is the arithmetic mean of all true targets. The numerator of Eq. (8) is the residual sum of squares (model error), and the denominator is the total sum of squares (data variance around the mean). Overall, these definitions allow rMSE and MAE to quantify average prediction error in the original units of $y$, whereas $R^2$ measures the proportion of variance in the data explained by the model.

V. EXPERIMENTAL RESULTS AND ANALYSIS

*A. Simulated soil GHG flux prediction*

Agricultural process-based simulation winter wheat dataset to evaluate the fertilisation-oriented soil GHG fluxes prediction performance of the proposed approach with the following comparative regression methods. For non-deep baseline methods, we select ridge [28], lasso [29] and elastic-net [30] regression. For baseline deep learning architectures, we select ResNet [31], GCN [32] and GraphSAGE [33]. Table II and III present the experimental results for soil $CO_2$ and $N_2O$ flux prediction respectively. Fig. 2 and 3 demonstrate the $R^2$ experimental results for soil $CO_2$ and $N_2O$ flux prediction respectively.

The following are our primary observations:
1) For the agricultural process-based simulation winter wheat dataset, the prediction accuracy of the proposed approach outperforms the comparative regression models, demonstrating the effectiveness of the agricultural feature decoupling extraction and structured correlation modelling between agricultural variables in our regression framework.

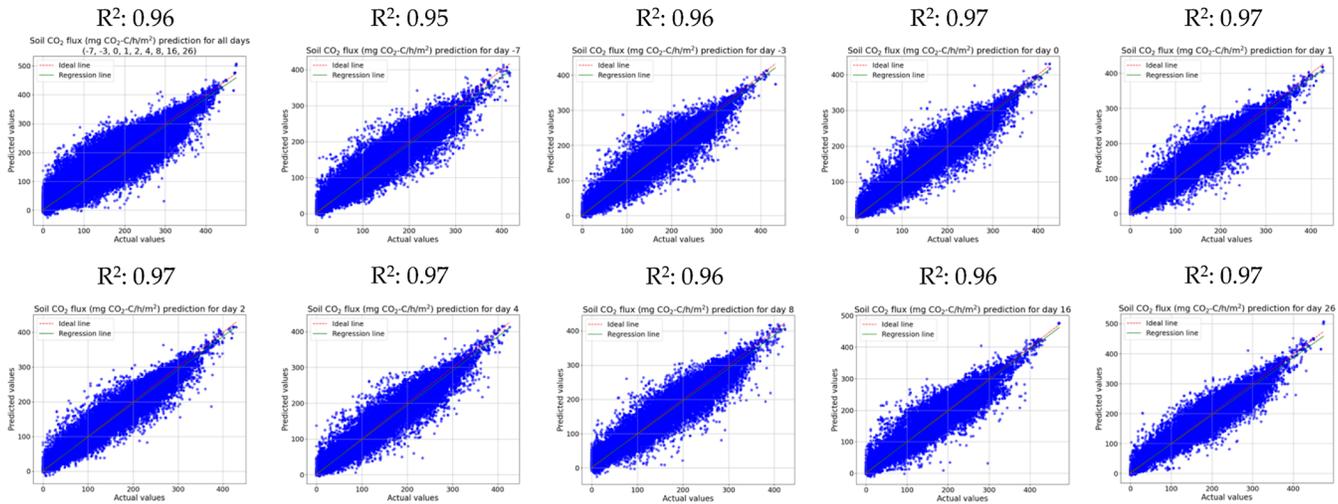

**Fig. 2.** Scatter plots of actual soil $CO_2$ flux values versus predicted values on agricultural process-based simulation testing data with our proposed approach. The red dash line in each figure is a reference of perfect prediction (Ideal line: predicted value exactly equals to actual value). We perform least squares regression on the points shown in the scatter plots and the green solid line is the regression line, which serves as a visual indicator of overall performance. The closer between the regression line and the ideal line, the better are the prediction results.

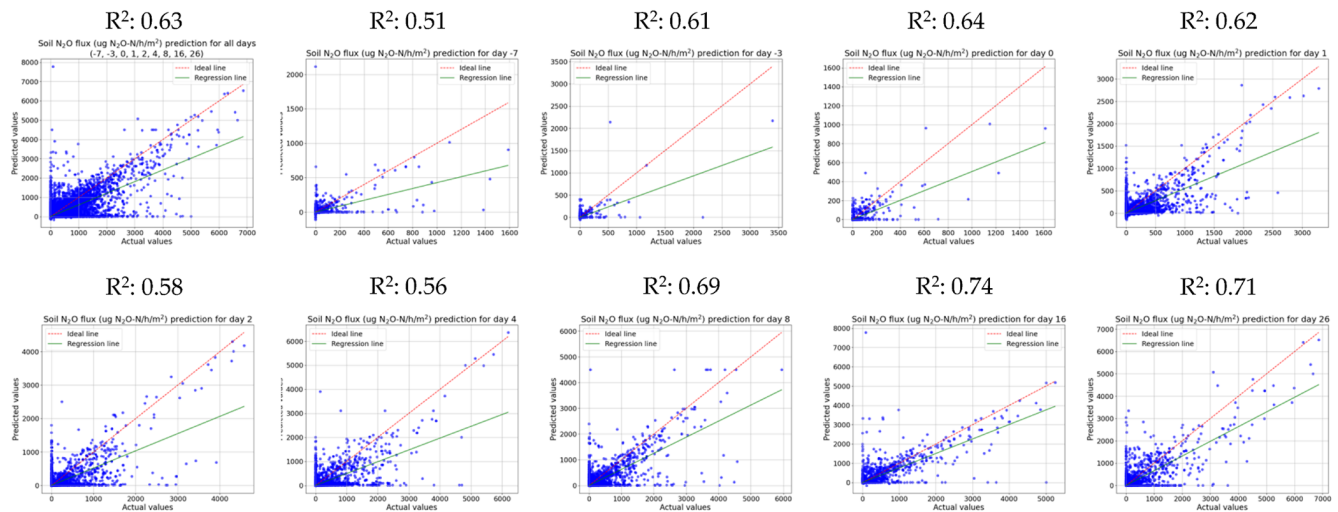

**Fig. 3.** Scatter plots of actual soil $N_2O$ flux values versus predicted values on agricultural process-based simulation testing data with our proposed approach. The red dash line in each figure is a reference of perfect prediction (Ideal line: predicted value exactly equals to actual value). We perform least squares regression on the points shown in the scatter plots and the green solid line is the regression line, which serves as a visual indicator of overall performance. The closer between the regression line and the ideal line, the better are the prediction results.

2) The proposed knowledge-guided GNN regression framework significantly enhances prediction stability. The standard deviation of the twenty iterations of the experiment was lower than that of the comparative regression models. This may be due to the structured correlation modelling between agricultural variables that enhances the stability and generalisation of the proposed approach, and the multi-target multi-graph structure enables the construction of individual agricultural knowledge correlation graphs for various soil GHG flux prediction targets, with each graph specialising in modelling important correlations at a particular time, effectively preventing the noise interference caused by mixing all the heterogeneous information in a single large graph, and enabling the model to precisely capture the factors influencing variations in soil GHG fluxes at different times or different days.

3) The $R^2$ test results demonstrate that the proposed approach achieves superior performance in predicting soil $CO_2$ fluxes, with $R^2$ exceeding 0.9 for both the overall and daily prediction. For the soil $N_2O$ flux prediction, the predictive performance of the proposed approach was marginally unstable, with an overall $R^2$ of 0.63, which can be attributed to the high spatial and temporal variability of $N_2O$ emissions, causing difficulties in precisely monitoring and quantifying the emissions.

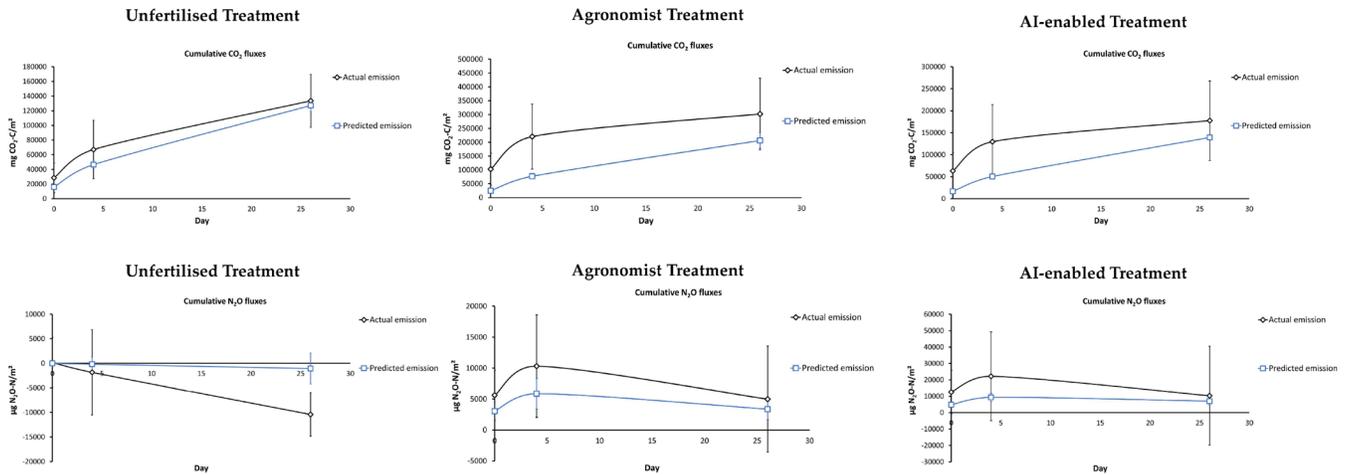

**Fig. 4.** The comparison of actual and predicted emissions for various fertiliser application plans on a genuine farm in the UK.

*B. Real-world soil GHG flux prediction*

To validate the application capability of our proposed approach in real-world scenarios, we conducted soil GHG flux predictions for three different fertilisation plans on a genuine farm in the UK. Three fertilisation plans were: 1) Unfertilised treatment, 2) Agronomist treatment, and 3) AI-enabled treatment. Figure 4 presents the comparison of actual and predicted emissions for various fertiliser application plans. The results demonstrate that our proposed method is currently unable to predict highly accurate soil GHG emission values at all time points, but it can certainly capture soil GHG emission trends.

## VI. CONCLUSION

In this research, we construct an agricultural process-based multi-dimensional simulation dataset covering 47 countries and propose a knowledge-guided GNN foundation model for fertilisation-oriented soil GHG flux prediction. The approach integrates the multi-dimensional agricultural simulation data generated by the agricultural process model DSSAT to enable precise modelling of soil GHG (including $CO_2$ and $N_2O$) fluxes for various fertiliser application conditions by constructing graph-structured interactions of heterogeneous agricultural features from multiple sources. Specifically, we utilised a modified autoencoder to extract crucial features from highly coupled agricultural multi-dimensional data, then design and construct a multi-target multi-graph GNN framework to model structured correlations between agricultural variables at various prediction targets and time points. The predictive model can be utilised to assess the impacts of fertilisation plans on soil GHG emissions and to identify coupling mechanisms between pivotal fertilisation practices and emission responses, thus supporting agricultural emission mitigation management and low-carbon decision-making. The experimental results demonstrate that the proposed approach outperforms the comparative methods in terms of accuracy and stability of soil GHG flux prediction with the agricultural process-based simulation dataset, and the proposed approach exhibits high stability in soil $CO_2$ flux prediction and certain volatility in soil $N_2O$ flux prediction, mainly due to its strong spatial and temporal variability. We further conducted empirical evaluations on real-world farm for three different fertilisation strategies (unfertilised treatment, agronomist treatment and AI-enabled treatment), and the results indicate that the model can effectively capture the overall trend of soil GHG emissions, although there is a certain level of bias in the absolute emission predictions at partial time points. Overall, the approach demonstrates superior generalisation ability and robustness under data-constrained scenarios, and is applicable to soil GHG emission modelling and decision support in complex agricultural management conditions.